\documentclass[twoside,11pt]{article}

\usepackage{blindtext}

\usepackage{jmlr2e}

\usepackage[numbers, compress]{natbib}

\usepackage{lastpage}

\ShortHeadings{Constrained Sliced Wasserstein Embedding}{NaderiAlizadeh, Salehi, Liu, and Kolouri}
\firstpageno{1}

\usepackage[utf8]{inputenc} 
\usepackage[T1]{fontenc}    
\usepackage{hyperref}       
\usepackage{url}            
\usepackage{booktabs}       
\usepackage{amsfonts}       
\usepackage{nicefrac}       
\usepackage{microtype}      
\usepackage{xcolor}         
\usepackage{amsmath,amsfonts}
\usepackage{mathtools}
\usepackage{dsfont}
\usepackage{color, colortbl}
\usepackage{multirow, multicol}
\usepackage{xparse}
\usepackage{tabularx}
\usepackage{lscape} 
\usepackage{hhline}
\definecolor{Gray}{gray}{0.9}
\usepackage{algorithm}
\usepackage{algpseudocode}
\usepackage{algcompatible}
\newcommand{\algcomment}[1]{\hfill// {\small\texttt{#1}}}

\usepackage[font=footnotesize]{caption}
\usepackage{subcaption}    
\usepackage{wrapfig}
\usepackage{enumitem}

\begin{document}

\title{Constrained Sliced Wasserstein Embedding}

\author{\name Navid NaderiAlizadeh \email navid.naderi@duke.edu \\
\addr Department of Biostatistics and Bioinformatics\\
Duke University\\
Durham, NC 27708, USA
\AND
\name Darian Salehi \email darian.salehi@duke.edu \\
\addr Department of Computer Science\\
Duke University\\
Durham, NC 27708, USA
\AND
\name Xinran Liu \email xinran.liu@vanderbilt.edu \\
\addr Department of Computer Science\\
Vanderbilt University\\
Nashville, TN 37212, USA
\AND
\name Soheil Kolouri \email soheil.kolouri@vanderbilt.edu \\
\addr Department of Computer Science\\
Vanderbilt University\\
Nashville, TN 37212, USA
}

\maketitle

\begin{abstract}
Sliced Wasserstein (SW) distances offer an efficient method for comparing high-dimensional probability measures by projecting them onto multiple 1-dimensional probability distributions. However, identifying informative slicing directions has proven challenging, often necessitating a large number of slices to achieve desirable performance and thereby increasing computational complexity. We introduce a constrained learning approach to optimize the slicing directions for SW distances. Specifically, we constrain the 1D transport plans to approximate the optimal plan in the original space, ensuring meaningful slicing directions. By leveraging continuous relaxations of these transport plans, we enable a gradient-based primal-dual approach to train the slicer parameters, alongside the remaining model parameters. We demonstrate how this constrained slicing approach can be applied to pool high-dimensional embeddings into fixed-length permutation-invariant representations. Numerical results on foundation models trained on images, point clouds, and protein sequences showcase the efficacy of the proposed constrained learning approach in learning more informative slicing directions. Our implementation code can be found at \url{https://github.com/Stranja572/constrainedswe}.
\end{abstract}


\section{Introduction}
Optimal Transport (OT) is a framework for finding the most efficient way to move one distribution of mass (or probability measure) to another, minimizing a specified cost associated with the transportation. It has a long-standing history in mathematics \cite{villani2008optimal} and continues to thrive as a vibrant field of study, seamlessly blending deep theoretical insights with practical applications. Recent advances in OT have garnered significant attention in the deep learning community across many domains such as computer vision \cite{bonneel2023survey, arjovsky2017wasserstein, korotin2019wasserstein, adrai2023deep, rout2021generative}, natural language processing \cite{huynh2020otlda, chen2023plot, li2024gilot}, medical imaging \cite{wang2024autoencoder, liu2024linear, oh2020unpaired}, and biology \cite{yang2020predicting, naderializadeh2025aggregating}. 
OT enables distribution alignment and provides metrics such as the Wasserstein distance, which can serve as effective loss functions in optimization tasks \cite{montesuma2024recent}. Moreover, OT has been used for data simplification methods that are essential for revealing the underlying structure in complex datasets, including clustering \cite{liu2023cot, laclau2017co, chakraborty2020hierarchical}, dimensionality reduction \cite{van2024distributional, meng2020sufficient}, and feature aggregation or pooling \cite{mialon2021trainable, kolouri2021wasserstein, naderializadeh2021PSWE}.

A prominent approach that empowers the application of OT in deep learning is linear OT \cite{wang2013linear, kolouri2016continuous} (also known as Wasserstein embedding). It acts as a measure-to-vector operator, allowing deep neural networks to handle measure-valued data without compromising the geometric structure. 
Linear OT embeddings have a fixed size and are invariant to permutations in the input distribution. This characteristic is particularly useful when developing permutation-invariant network structures \cite{zaheer2017deep, lee2019set} for inherently unordered data types, such as point clouds \cite{lu2024slosh}, graph node embeddings~\cite{kolouri2021wasserstein,nikolentzos2017matching}, or features extracted from images \cite{dosovitskiy2020image}. 
In these neural networks, a pooling layer is typically inserted after an equivariant backbone to aggregate the extracted features, which helps reduce the complexity and mitigate overfitting. Pooling mechanisms, such as mean, sum, and max operators, need to provide the network with specific invariances, such as translation \cite{zafar2022comparison} or permutation invariance \cite{zaheer2017deep}. Thus, linear OT can play a crucial role in pooling due to its permutation-invariant nature along with its strong ability to capture geometric structure.

Despite its advantages, OT is often hindered by high computational costs. Standard solvers for discrete OT problems leverage linear programming, typically resulting in a computational complexity of $\mathcal{O}(M^3\log M)$ when dealing with distributions supported on $M$ discrete points~\cite{peyre2019computational}. 
Among the proposed alternatives~\cite{cuturi2013sinkhorn}, sliced OT~\cite{bonneel2015sliced} improves efficiency by projecting high-dimensional distributions onto 1-dimensional slices, where a closed-form solution exists. The resulting sliced Wasserstein (SW) distance can be computed with a time complexity of $\mathcal{O}(LM\log M)$ for $L$ slices. Linear Optimal Transport embeddings can be extended to the sliced OT framework by computing Wasserstein embeddings for each individual slice, which yields the sliced Wasserstein embedding (SWE) \cite{kolouri2015radon, naderializadeh2021PSWE,shifat2021radon}. SWE inherits the permutation-invariant property of the Wasserstein embedding while offering enhanced computational efficiency, making it a strong candidate for the pooling layer in more complex network architectures.

The computational efficiency of the slicing approach, however, comes at the cost of projection complexity \cite{nadjahi2020statistical}. A large number of slices is often needed to accurately capture dissimilarities between distributions, particularly in high-dimensional spaces. This challenge has motivated research on identifying the most informative slices. Some approaches measure the importance of a slice based on how well it distinguishes between the projected distributions~\cite{deshpande2019max,dai2021sliced,nguyen2021distributional,tran2024understanding}, while others use non-linear projections to better capture the complex structures of high-dimensional data \cite{kolouri2019generalized, chen2020augmented}.


In this work, we propose a constrained learning framework to optimize the slicing directions in Sliced Wasserstein Embeddings (SWE). Specifically, we constrain the transport plans obtained from the slices to approximate the optimal transport plan in the original high-dimensional space. Our approach is motivated by recent advances in sliced Wasserstein generalized geodesics (SWGG)~\cite{mahey2023fast} and expected sliced transport plans~\cite{liu2025expected}. We focus specifically on evaluating the effectiveness of this constrained SWE as a pooling method, where we leverage a primal-dual training algorithm to find the right balance between minimizing a primary objective function and satisfying the aforementioned constraints on the transport plans. Our contributions are as follows:

\begin{itemize}[leftmargin=*]
    \item We propose a novel constrained learning framework imposing SWGG dissimilarity constraints on the slicing directions, with automatic constraint relaxation, as needed, to ensure feasibility.
    \item We develop a primal-dual training algorithm to solve this constrained learning problem in the context of pooling high-dimensional embeddings via continuous relaxations of permutation matrices.
    \item We empirically show that our proposed constrained embeddings enhance the downstream performance of pre-trained foundation models on images, point clouds, and protein sequences.
\end{itemize}

\section{Background and Related Work}

\subsection{Wasserstein Distances}

Wasserstein distances arise from the optimal mass transportation problem, where one is interested in finding a transportation plan (between two distributions) that leads to the least expected transportation cost for a given ground metric (or transportation cost)~\cite{villani2008optimal}. Consider two probability measures $\mu$ and $\nu$ with finite 2\textsuperscript{nd} moments defined on $\mathbb{R}^d$. Let $\Gamma(\mu,\nu)$ denote the set of all transportation plans $\gamma$ such that $\gamma(A\times\mathbb{R}^d)=\mu(A)$ and $\gamma(\mathbb{R}^d \times A)=\nu(A)$ for any measurable set $A\subseteq \mathbb{R}^d$. Then, the (2-) distance between $\mu$ and $\nu$ is defined as
\begin{align}\label{eq:w2}
\mathcal{W}_2(\mu,\nu) \coloneqq \left(\inf_{\gamma\in\Gamma(\mu,\nu)} \int_{\mathbb{R}^d \times \mathbb{R}^d} \|\mathbf{x} - \mathbf{y}\|_2^2 ~ d\gamma(\mathbf{x} , \mathbf{y})\right)^{\frac{1}{2}}.
\end{align}
The plan $\gamma^*\in\Gamma(\mu,\nu)$ that is the solution to the minimization problem in~\eqref{eq:w2} is called the optimal transport (OT) plan, and it represents how to move the probability mass from $\mu$ to $\nu$ with the lowest possible $\ell_2$ cost. Wasserstein distances have gained significant interest in a wide variety of areas in machine learning as geometric-aware distances to compare distributions. For instance, these distances have been used in applications concerning domain adaptation~\cite{courty2016optimal}, transfer learning \cite{alvarez2020geometric}, generative learning \cite{nguyen2021distributional}, reinforcement learning~\cite{zhang18a, moskovitz2021efficient}, and imitation learning~\cite{xiao2019wasserstein,zhang2020nested,dadashi2021primal}.

Despite the utility and desirable properties of Wasserstein distances, calculating them in practice incurs a high computational complexity. For two empirical distributions, each with $M$ samples, the computational complexity of solving the minimization problem in~\eqref{eq:w2} is $\mathcal{O}(M^3\log M)$~\cite{bonneel2011displacement,peyre2019computational}. Entropy-regularized approximation of OT reduces the computational complexity to $\mathcal{O}(M^2)$~\cite{cuturi2013sinkhorn}. A notable exception is the case of one-dimensional probability measures, where the computational complexity dramatically drops to $\mathcal{O}(M\log M)$. This is thanks to a closed-form solution based on sorting and matching that solves~\eqref{eq:w2}. Letting $F_{\mu}$ and $F_{\nu}$ denote the cumulative distribution functions (CDFs) of $\mu$ and $\nu$ defined on $\mathbb{R}$, the 2-Wasserstein distance between them can be written as~\cite{rabin2012wasserstein}
\begin{align}\label{eq:w2_1d}
    \mathcal{W}_2(\mu,\nu)=\left(\int_{0}^1 \left\|F^{-1}_{\mu}(t)-F^{-1}_{\nu}(t)\right\|_2^2 dt\right)^{\frac{1}{2}}.
\end{align}

As the inverse of the CDF is also referred to as the quantile function, the Wasserstein distance between two one-dimensional distributions is equivalent to the Euclidean distance between their corresponding quantile functions. This closed-form solution for one-dimensional measures has led to a line of research on sliced Wasserstein distances, which we review next.

\subsection{Sliced Wasserstein Distances}

The key idea behind sliced OT and sliced Wasserstein (SW) distances is that high-dimensional distributions can be projected onto several one-dimensional \emph{slices}, in each of which the Wasserstein distance has a closed-form solution~\eqref{eq:w2_1d}~\cite{bonneel2015sliced,kolouri2016sliced, kolouri2018sliced, deshpande2019max, kolouri2019generalized, zhang2023projection}. In particular, letting $\mathbb{S}^{d-1}$ denote the unit hypersphere in $\mathbb{R}^d$, the SW distance between $\mu$ and $\nu$ is defined as the expected Wasserstein distance among all projections $\theta \in \mathbb{S}^{d-1}$ of $\mu$ and $\nu$, i.e.,
\begin{align}\label{eq:swd}
\mathcal{SW}_2(\mu,\nu) \coloneqq \left( \int_{\mathbb{S}^{d-1}} \mathcal{W}_2^2 \left( \theta_{\#} \mu, \theta_{\#} \nu \right) ~ d\theta \right)^{\frac{1}{2}},
\end{align}
where $\theta_{\#} \mu$ and $\theta_{\#} \nu$ denote the corresponding one-dimensional projected measures onto $\theta$. In practice, since calculating the integration in~\eqref{eq:swd} across an infinite number of slices is infeasible, we resort to an empirical approximation across a set of $L$ slices,
\begin{align}\label{eq:swd_finite_slices}
\mathcal{SW}_2(\mu,\nu) \approx \left( \sum_{l=1}^L \sigma_l \mathcal{W}_2^2 \left( \theta_{l\#} \mu, \theta_{l\#} \nu \right) \right)^{\frac{1}{2}},
\end{align}
where $\sigma_l \geq 0, \forall l\in\{1,\dots,L\}$ and $\sum_{l=1}^L \sigma_l = 1$. The quality of the approximation in~\eqref{eq:swd_finite_slices} depends on both the number and the ``quality'' of slices. In particular, for large $d$, the number of slices, $L$, typically needs to be very large, which proportionally increases the computation complexity. Prior studies mainly focus either on finding a single, maximally informative slice~\cite{deshpande2019max,kolouri2019generalized,mahey2023fast} or sampling a larger number of slices in an effective manner~\cite{paty2019subspace,nadjahi2021fast,nguyen2021distributional,nguyen2023hierarchical,nguyen2023energybased}. For example, the Max-SW \cite{deshpande2019max} (and Max$K$-SW \cite{dai2021sliced}) utilize a single slice (or $K$ slices) that induces the largest (or top-$K$) projected distances. Distributional-SW \cite{nguyen2021distributional} identifies an optimal distribution of slices on which the expectation of 1-dimensional Wasserstein distances is maximized, whereas Markovian SW \cite{nguyen2023markovian} finds an optimal Markov chain of slices. Energy-Based SW \cite{nguyen2023energybased} assigns greater weight to the slices with higher values of a monotonically increasing energy function of the projected distance.

Of particular relevance to this work is the notion of sliced Wasserstein generalized geodesics (SWGG) \cite{mahey2023fast}. Consider two discrete probability measures $\mu = \sum_{\mathbf{x} \in \mathbb{R}^d} p(\mathbf{x}) \delta_{\mathbf{x}}$ and $\nu = \sum_{\mathbf{y} \in \mathbb{R}^d} q(\mathbf{y}) \delta_{\mathbf{y}}$ in $\mathcal{P}(\mathbb{R}^d)$. Given a slicing direction $\theta \in \mathbb{S}^{d-1}$, there exists a unique OT plan between the sliced distributions $\theta_{\#} \mu$ and $\theta_{\#} \nu$, denoted by $\Lambda_{\theta}^{\mu, \nu}$. Leveraging the quotient space of these 1D distributions~\cite{liu2025expected,shahbazi2025espformer_arxiv}, we can construct a lifted transport plan in the original $d$-dimensional space, given by
\begin{align}\label{eq:lifted_plan}
\gamma_\theta^{\mu, \nu} \coloneqq \sum_{\mathbf{x}\in\mathbb{R}^d}\sum_{\mathbf{y}\in\mathbb{R}^d}u_\theta^{\mu, \nu}(\mathbf{x}, \mathbf{y})\delta(\mathbf{x}, \mathbf{y}),
\end{align}
where $u_\theta^{\mu, \nu}$ is defined as
\begin{align}
u_\theta^{\mu, \nu}(\mathbf{x}, \mathbf{y}) \coloneqq
\frac{p(\mathbf{x})q(\mathbf{y})}{P_{\theta}(\mathbf{x}) Q_{\theta}(\mathbf{y})}\Lambda_\theta^{\mu, \nu}(\theta^T \mathbf{x}, \theta^T \mathbf{y}), \quad \forall \mathbf{x},\mathbf{y}\in\mathbb{R}^d,
\end{align}
and $P_{\theta}(\mathbf{x})$ and $Q_{\theta}(\mathbf{y})$ are defined as
\begin{align}
P_{\theta}(\mathbf{x}) \coloneqq \sum_{\mathbf{x}'\in\mathbb{R}^d:\theta^T \mathbf{x}'=\theta^T \mathbf{x}}p(\mathbf{x}'), \quad Q_{\theta}(\mathbf{y}) \coloneqq \sum_{\mathbf{y}'\in\mathbb{R}^d:\theta^T \mathbf{y}'=\theta^T \mathbf{y}}q(\mathbf{y}'),  \quad \forall \mathbf{x},\mathbf{y}\in\mathbb{R}^d.
\end{align}

Having the lifted transport plan in~\eqref{eq:lifted_plan}, we can then write the SWGG metric~\cite{mahey2023fast,liu2025expected} as
\begin{align}\label{eq:swgg}
\mathcal{D}_2(\mu, \nu; \theta) = \left(\sum_{\mathbf{x}\in\mathbb{R}^d}\sum_{\mathbf{y}\in\mathbb{R}^d}\|\mathbf{x}-\mathbf{y}\|_2^2 \gamma_{\theta}^{\mu, \nu}(\mathbf{x}, \mathbf{y})\right)^\frac{1}{2}.
\end{align}
Importantly, we have $\mathcal{W}_2(\mu,\nu)\leq \mathcal{D}_2(\mu,\nu;\theta)$ for $\forall \theta\in \mathbb{S}^{d-1}$. Hence, min-SWGG \cite{mahey2023fast} proposes to minimize the upper bound with respect to $\theta$ to obtain (nearly) optimal transportation plans.

\subsection{Representation Learning using Wasserstein Distances}

Wasserstein distances have been used for representation learning from unstructured data, e.g., graphs and sets \cite{togninalli2019wasserstein,mialon2021a,kolouri2021wasserstein,naderializadeh2021PSWE,haviv2024wasserstein}. The core idea behind these approaches is to treat a set of $d$-dimensional features (e.g., node-embeddings of a graph) as an empirical distribution and compare various sets via their Wasserstein distance or the variations of this distance, e.g., sliced Wasserstein distance.
To reduce the computational overhead for pairwise comparison of such empirical distributions, recent work leverages Wasserstein embeddings \cite{kolouri2021wasserstein,haviv2024wasserstein} which map the input sets (i.e., the empirical distributions) into a vector space, in which the Euclidean distance approximates the Wasserstein distance between the input distributions. Sliced Wasserstein embeddings have also been investigated as pooling operators following permutation-equivariant backbones
~\cite{naderializadeh2021PSWE, kothapalli2024equivariant, amir2025fourier}.

\subsection{Learning under Constraints}

Traditional problems in machine learning 
are typically formulated as \emph{unconstrained} optimization problems, where an objective function of interest is minimized (e.g., in the case of a loss function) or maximized (e.g., in the case of a reward function). However, in many application domains, such as autonomous driving~\cite{lefevre2015learning,zhang2023adaptive,gao2024constraints}, robotics~\cite{chou2018learning,marco2021robot,liu2022robot}, networking~\cite{eisen2019learning,shi2023machine,naderializadeh2023learning}, and healthcare~\cite{hakim2024need,gangavarapu2024enhancing,yang2024ensuring}, there are certain requirements, constraints, or guardrails that the learning-based systems need to respect. Moreover, in some scenarios, certain characteristics are desired from a machine learning model to make it more generalizable, such as reduced magnitude of model parameters to mitigate overfitting~\cite{salman2019overfitting,kolluri2020reducing,santos2022avoiding}, or increased action distribution entropy in reinforcement learning to promote action diversity~\cite{o'donoghue2017combining,levine2018reinforcement,zhao2019maximum,adamczyk2023utilizing,massiani2024viability}.

The most common approach to training machine learning models that consider and satisfy such constraints is by modifying the primary objective to promote the constraints using fixed coefficients, an approach referred to as regularization~\cite{moradi2020survey,tian2022comprehensive,kotsilieris2022regularization}. However, such regularization-based techniques come with an increased complexity of tuning the regularization coefficient~\cite{gallego2022controlled}. Moreover, they do not come with any theoretical guarantees about achieving certain desired bounds on the constraints that the model is attempting to satisfy. A more principled way of addressing such requirements is \emph{constrained learning}, where the learning problem is reformulated as a constrained optimization problem~\cite{chamon2020probably,hounie2023resilient,bai2022achieving,hong2024primal}. This way, the model attempts to strike the right trade-off between optimizing the objective and satisfying the requirements posed on the model~\cite{chamon2022constrained,gallego2022controlled,calvo2023state,ramirez2025feasible,elenter2022lagrangian,elenter2023primal}.

\section{Proposed Method}

Consider a generic minimization problem over a set of $L$ slices $\boldsymbol{\Theta}=[\theta_1 ~ \dots ~ \theta_L]^T \in (\mathbb{S}^{d-1})^L$ for a fixed pair of probability measures $\mu$ and $\nu$, formulated as
\begin{align}\label{eq:opt_unconstrained}
\min_{\boldsymbol{\Theta} \in (\mathbb{S}^{d-1})^L} f(\boldsymbol{\Theta}; \mu, \nu),
\end{align}
where $f(\cdot; \mu, \nu):(\mathbb{S}^{d-1})^L \to \mathbb{R}$ denotes the objective/loss function conditioned on $\mu$ and $\nu$. Motivated by~\cite{mahey2023fast}, we hypothesize that a ``good'' slice $\theta \in \mathbb{S}^{d-1}$ is one for which the SWGG dissimilarity $\mathcal{D}_2(\mu, \nu; \theta)$ in~\eqref{eq:swgg} is as small as possible, meaning that the sliced transport plan is also highly relevant in the original space. We impose this notion as SWGG \emph{constraints} in the optimization problem, where we reformulate the unconstrained problem in~\eqref{eq:opt_unconstrained} as a \emph{constrained learning} problem,
\begin{subequations}\label{eq:opt_constrained}
\begin{alignat}{2}
&\min_{\boldsymbol{\Theta} \in (\mathbb{S}^{d-1})^L}  &~~& f(\boldsymbol{\Theta}; \mu, \nu), \\
&\qquad\text{s.t.} && \mathcal{D}_2(\mu, \nu; \theta_l) \leq \epsilon_l, \quad \forall l\in\{1,\dots,L\}. \label{eq:swgg_constraint}
\end{alignat}
\end{subequations}

\setlength{\columnsep}{16pt}%
\begin{wrapfigure}[19]{r}{0.60\linewidth}
\vspace{-.07in}
\captionsetup{belowskip=0pt}
\includegraphics[width=\linewidth]{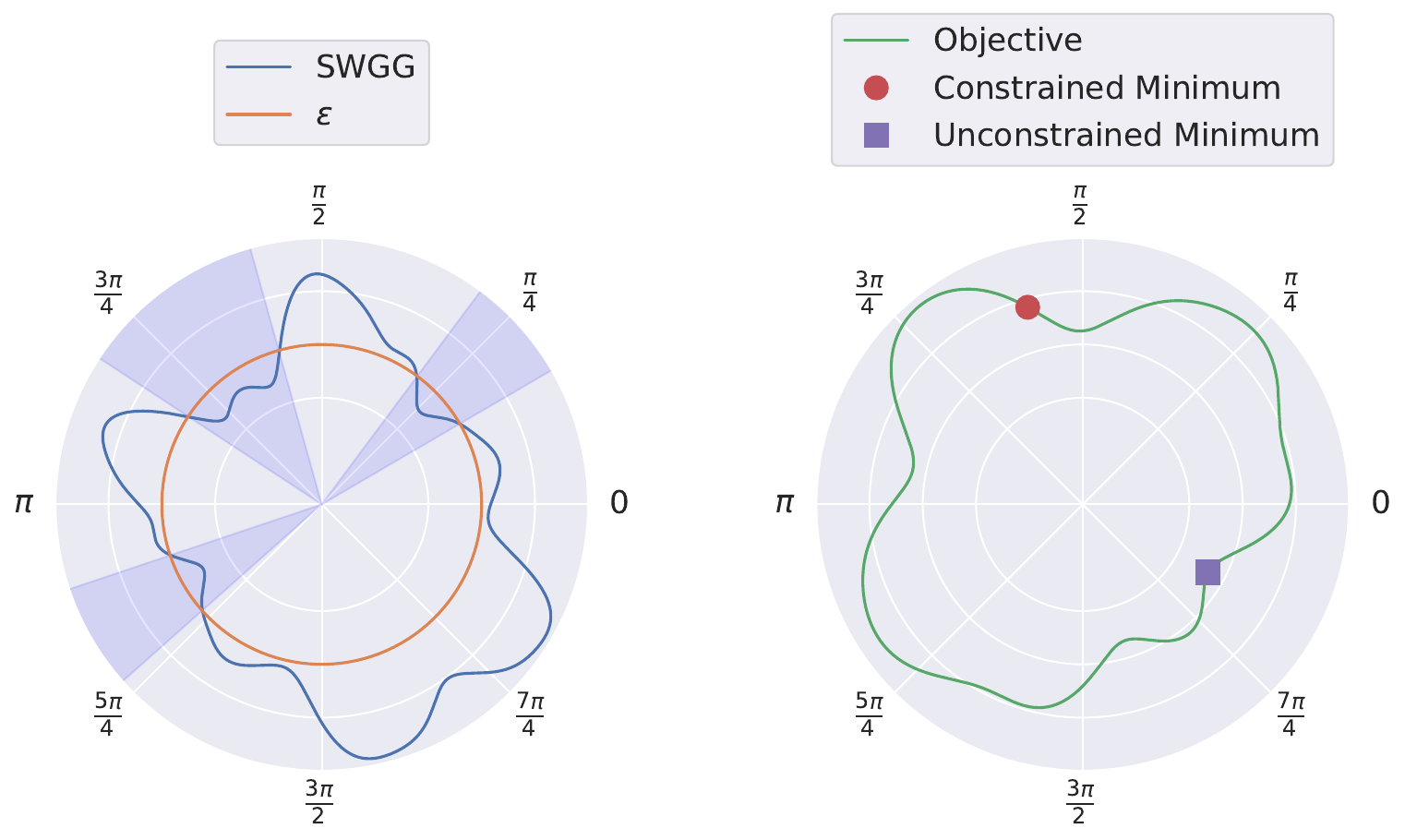}%
\caption{(Left) Example of SWGG values for a pair of distributions in $\mathbb{R}^2$, and (Right) the objective function values for a single slice ($L=1$). Our proposed method enforces requirements on the slicing directions, requiring the optimization problem to focus on a feasible subset of slices whose SWGG values are bounded by a constant (the shaded areas in the left plot), hence impacting the final solution.}%
\label{fig:swgg_obj_teaser}
\end{wrapfigure}
~\hspace{-.05in}
In~\eqref{eq:swgg_constraint}, $\epsilon_l$ denotes the SWGG dissimilarity upper bound enforced on the $l$\textsuperscript{th} slice, $l\in\{1,\dots,L\}$. As shown in Figure~\ref{fig:swgg_obj_teaser}, the proposed constrained learning formulation in~\eqref{eq:opt_constrained} restricts the search space for the slicing directions, guiding the learning problem to select slices that strike the right trade-off between minimizing the primary objective function and respecting the SWGG dissimilarity upper bounds.

The feasibility of the problem~\eqref{eq:opt_constrained} crucially depends on the choice of the upper bounds $\{\epsilon_l\}_{l=1}^L$---extremely small values render the problem~\eqref{eq:opt_constrained} infeasible. On the other hand, the problem reverts to the original unconstrained formulation in~\eqref{eq:opt_unconstrained} if large values are assigned to these upper bounds. We propose to use the notion of \emph{resilience} in constrained learning~\cite{hounie2023resilient} to slightly relax the constrained problem~\eqref{eq:opt_constrained}, as needed, to make it feasible. In particular, we introduce non-negative \emph{slack} variables $\mathbf{s}=[s_1 ~ \dots ~ s_L]^T\in\mathbb{R}_+^L$, which are used in the relaxed formulation of~\eqref{eq:opt_constrained} as follows,\vspace{-.03in}
\begin{subequations}\label{eq:opt_constrained_relaxed}
\begin{alignat}{2}
&\min_{\boldsymbol{\Theta} \in (\mathbb{S}^{d-1})^L,~\mathbf{s}\in\mathbb{R}_+^L} &~~& f(\boldsymbol{\Theta}; \mu, \nu) + \frac{\alpha}{2} \|\mathbf{s}\|_2^2, \label{eq:obj_relaxed}        \\
&\qquad\quad\text{s.t.} && \mathcal{D}_2(\mu, \nu; \theta_l) \leq \epsilon_l + s_l, \quad \forall l\in\{1,\dots,L\}, \label{eq:swgg_constraint_relaxed}
\end{alignat}
\end{subequations}
where in~\eqref{eq:obj_relaxed}, $\alpha \geq 0$ denotes a small non-negative coefficient that prevents the slack variables from growing too large. The formulation in~\eqref{eq:opt_constrained_relaxed}, in effect, relaxes the original problem~\eqref{eq:opt_constrained} just enough to find a feasible set of slicers that satisfy the (relaxed) SWGG dissimilarity requirements.

\begin{remark}
Even though the learning problems in~\eqref{eq:opt_unconstrained}-\eqref{eq:opt_constrained_relaxed} are formulated for a single pair of probability measures $\mu$ and $\nu$, it is straightforward to extend them to multiple measures  $\{\mu_i\}_{i=1}^N$ and $\{\nu_i\}_{i=1}^N$. We present one such example in Section~\ref{sec:cswe}.
\end{remark}

\begin{remark}
In this paper, we focus on SWGG-based constraints for enhancing the informativeness of the optimized slicing directions. However, our constrained learning formulation can also include additional types of constraints in the optimization problem, such as orthogonality constraints on the slices or lower bounds on the sliced Wasserstein distances~\cite{deshpande2019max}. In our experiments, we observed minimal differences by adding orthogonality constraints (in particular, $\|\boldsymbol{\Theta}^T \boldsymbol{\Theta}-\mathbb{I}_L\|\leq \delta$, where $\mathbb{I}_L$ denotes the $L \times L$ identity matrix), but the utility of the constraints could depend on the task under study. We leave the extension of the proposed method to other constraint types and their evaluation in tasks beyond those studied in Section~\ref{sec:experiments} as future work.
\end{remark}

\subsection{Primal-Dual Constrained Learning of Slices}\vspace{-.03in}

To solve the relaxed problem~\eqref{eq:opt_constrained_relaxed}, we move to the Lagrangian dual domain~\cite{boyd2004convex,fioretto2021lagrangian}. In particular, we assign a set of non-negative \emph{dual variables} $\boldsymbol{\lambda} =[\lambda_1 ~ \dots ~ \lambda_L]^T\in\mathbb{R}_+^L$ to the constraints~\eqref{eq:swgg_constraint_relaxed}, allowing us to write the \emph{Lagrangian} associated with~\eqref{eq:opt_constrained_relaxed} as
\vspace{-.03in}
\begin{align}
\mathcal{L}(\boldsymbol{\Theta}, \mathbf{s}, \boldsymbol{\lambda}) &= f(\boldsymbol{\Theta}; \mu, \nu) + \frac{\alpha}{2} \|\mathbf{s}\|_2^2 + \sum_{l=1}^L \lambda_l \Big[\mathcal{D}_2(\mu, \nu; \theta_l) - (\epsilon_l + s_l)\Big] \nonumber \\
&= f(\boldsymbol{\Theta}; \mu, \nu) + \frac{\alpha}{2} \|\mathbf{s}\|_2^2 + \boldsymbol{\lambda}^T \big[\boldsymbol{\mathcal{D}}(\boldsymbol{\Theta}) - (\boldsymbol{\epsilon} + \mathbf{s})\big],\label{eq:lagrangian}
\end{align}
where $\boldsymbol{\mathcal{D}}(\boldsymbol{\Theta}) \coloneqq [\mathcal{D}_2(\mu, \nu; \theta_1) ~ \dots ~ \mathcal{D}_2(\mu, \nu; \theta_L)]^T\in\mathbb{R}_+^L$ and $\boldsymbol{\epsilon} =[\epsilon_1 ~ \dots ~ \epsilon_L]^T\in\mathbb{R}_+^L$ represent the vectors of per-slice SWGG dissimilarities and upper bounds, respectively. Having the Lagrangian, we then formulate the dual problem of~\eqref{eq:opt_constrained_relaxed} as
\vspace{-.03in}
\begin{align}\label{eq:dual_problem}
\max_{\boldsymbol{\lambda}\in\mathbb{R}_+^L} ~ \min_{\boldsymbol{\Theta} \in (\mathbb{S}^{d-1})^L,~\mathbf{s}\in\mathbb{R}_+^L} \mathcal{L}(\boldsymbol{\Theta}, \mathbf{s}, \boldsymbol{\lambda}).
\end{align}
We then use a primal-dual approach to solve the dual problem~\cite{chamon2020probably,chamon2022constrained}, alternating between (projected) gradient descent steps on the primal variables  $\boldsymbol{\Theta}, \mathbf{s}$, i.e.,
\begin{align}
\boldsymbol{\Theta} &\leftarrow \boldsymbol{\Theta} -\eta_{\boldsymbol{\Theta}} \frac{\partial \mathcal{L}(\boldsymbol{\Theta}, \mathbf{s}, \boldsymbol{\lambda})}{\partial \boldsymbol{\Theta}}, \label{eq:slice_update} \\
\mathbf{s} &\leftarrow \left[\mathbf{s} - \eta_{\mathbf{s}} \frac{\partial \mathcal{L}(\boldsymbol{\Theta}, \mathbf{s}, \boldsymbol{\lambda})}{\partial \mathbf{s}}\right]_+ = \left[\mathbf{s} -\eta_{\mathbf{s}} (\alpha \mathbf{s} - \boldsymbol{\lambda})\right]_+,
\end{align}
and projected gradient ascent steps on the dual variables $\boldsymbol{\lambda}$, i.e.,
\begin{align}\label{eq:dual_updates}
\boldsymbol{\lambda} &\leftarrow \left[\boldsymbol{\lambda} + \eta_{\boldsymbol{\lambda}} \frac{\partial \mathcal{L}(\boldsymbol{\Theta}, \mathbf{s}, \boldsymbol{\lambda})}{\partial \boldsymbol{\lambda}}\right]_+ = \Big[\boldsymbol{\lambda}  + \eta_{\boldsymbol{\lambda}} \big[\boldsymbol{\mathcal{D}}(\boldsymbol{\Theta}) - (\boldsymbol{\epsilon} + \mathbf{s})\big]\Big]_+,
\end{align}
where $[\cdot]_+ \coloneqq \max(\cdot,0)$ represents projection onto the non-negative orthant, and $\eta_{\boldsymbol{\Theta}}$, $\eta_{\mathbf{s}}$, and $\eta_{\boldsymbol{\lambda}}$ denote the learning rates corresponding to the slice parameters, slack variables, and dual variables, respectively. An overview of the primal-dual algorithm is illustrated in Algorithm~\ref{alg:pd}.

It is important to note that the gradient ascent updates on the dual variables in~\eqref{eq:dual_updates} imply that the dual variable corresponding to each slice tracks how much that slice is violating its (relaxed) SWGG constraint: The higher the SWGG dissimilarity for a given slice, the larger its corresponding dual variable, and vice versa. This implies that the proposed primal-dual method of solving the dual problem~\eqref{eq:dual_problem} amounts to an adaptive regularization of the objective using the dual variables as dynamic regularization coefficients in~\eqref{eq:lagrangian}.

\setlength{\textfloatsep}{5pt}
\begin{algorithm}[t]
\caption{Primal-dual constrained learning of slices with (relaxed) SWGG upper bounds}
    \label{alg:pd}
    \begin{algorithmic}[1]
    \STATE {\bfseries Input:} Primal learning rate $\eta_{\boldsymbol{\Theta}}$, slack learning rate $\eta_{\mathbf{s}}$, dual learning rate $\eta_{\boldsymbol{\lambda}}$, slack regularization coefficient $\alpha$, constraint vector $\boldsymbol{\epsilon}$, and number of primal-dual iterations $T$.
    \STATE Initialize: $\boldsymbol{\Theta}$, $\boldsymbol{\lambda}\leftarrow\mathbf{0}$, $\mathbf{s}\leftarrow\mathbf{0}$.
    \FOR{$t=1, \ldots, T$}
        \STATE 
        $\boldsymbol{\Theta} \leftarrow \boldsymbol{\Theta} -\eta_{\boldsymbol{\Theta}} \frac{\partial \mathcal{L}(\boldsymbol{\Theta}, \mathbf{s}, \boldsymbol{\lambda})}{\partial \boldsymbol{\Theta}}
        $ \algcomment{Update slicer parameters}
        \STATE
        $
        \mathbf{s} \leftarrow \big[\mathbf{s} -\eta_{\mathbf{s}} (\alpha \mathbf{s} - \boldsymbol{\lambda})\big]_+
        $ \algcomment{Update slack variables}
        \STATE
        $
        \boldsymbol{\lambda} \leftarrow  \big[\boldsymbol{\lambda}  + \eta_{\boldsymbol{\lambda}} \big[\boldsymbol{\mathcal{D}}(\boldsymbol{\Theta}) - (\boldsymbol{\epsilon} + \mathbf{s})\big]\big]_+,
        $ \algcomment{Update dual variables}
    \ENDFOR
    \STATE {\bfseries Return:}{  $\boldsymbol{\Theta}$, $\boldsymbol{\lambda}$, $\mathbf{s}$.} 
    \end{algorithmic}
\end{algorithm}

\section{Case Study: Constrained Sliced Wasserstein Embeddings}\label{sec:cswe}
\vspace{-.03in}

As a use case of the proposed constrained learning method, we focus on sliced Wasserstein embedding (SWE)~\cite{naderializadeh2021PSWE}. This method leverages sliced optimal transport to derive permutation-invariant pooling by calculating the Monge coupling between the sliced empirical distributions corresponding to the input set of vectors and a trainable set of reference vectors.

Consider a supervised learning problem over a training dataset $\{(X_i, y_i)_{i=1}^N\}$, where $y_i\in\mathcal{Y}$ denotes the classification/regression target corresponding to the $i$\textsuperscript{th} training sample, and $X_i \in \mathcal{X}^{M_i}$ denotes the input corresponding to the $i$\textsuperscript{th} training sample, containing $M_i\in\mathbb{N}$ tokens. Example inputs could include point clouds, sequences, graphs, and images, whose size could vary across different samples. We consider a domain-specific, size-invariant backbone $g(\cdot; \phi): \mathcal{X}^{m} \to \mathbb{R}^{d \times m}, \forall m\in\mathbb{N}$, parameterized by $\phi\in\Phi$, that processes any given set of input tokens to a set of $d$-dimensional token-level embeddings, e.g., a transformer network.

Let $\mathbf{v}_{ij}\in\mathbb{R}^d$ represent the $j$\textsuperscript{th} token-level embedding of the $i$\textsuperscript{th} training sample, $i\in\{1,\dots,N\}, j\in\{1,\dots,M_i\}$, i.e., $g(X_i; \phi) = \mathbf{V}_i = [\mathbf{v}_{i1} ~ \dots ~ \mathbf{v}_{iM_i}]$. For a constant $M \in \mathbb{N}$, we consider a set of $M$ trainable \emph{reference} embeddings $\mathbf{U}=[\mathbf{u}_1 ~ \dots ~ \mathbf{u}_{M}]\in\mathbb{R}^{d \times M}$. For a given slice $\theta_l \in \mathbb{S}^{d-1}, l\in\{1,\dots,L\}$, the sliced empirical distributions induced by the reference set and the $i$\textsuperscript{th} training sample are given by $\mu^{l}=\frac{1}{M}\sum_{j=1}^{M} \delta_{\theta_l^T \mathbf{u}_{j}}$ and $\nu_i^{l}=\frac{1}{M_i}\sum_{j=1}^{M_i} \delta_{\theta_l^T \mathbf{v}_{ij}}$, respectively. Then, the Monge coupling between these two distributions is derived as part of the final aggregated embedding. In what follows, we describe the procedure when the number of tokens in any given input sample is fixed and equal to the number of tokens in the reference set, i.e., $M_i=M, \forall i\in\{1,\dots,N\}$. Details on how to extend the derivations for arbitrary set sizes are deferred to Appendix~\ref{appx:interp}.

Let $\mathbb{S}_M$ denote the set of all permutation matrices of order $M$. Sorting the $l$\textsuperscript{th} projected embeddings leads to two permutation matrices $\mathbf{P}^l, \mathbf{Q}_i^l \in \mathbb{S}_M$ that sort the projected reference and input embeddings in ascending order, respectively, i.e.,
\vspace{-.03in}
\begin{align}\label{eq:sort_perm}
(\theta_l^T \mathbf{U} \mathbf{P}^l)_1 \leq \dots \leq (\theta_l^T \mathbf{U} \mathbf{P}^l)_M, \qquad (\theta_l^T \mathbf{V}_i \mathbf{Q}_i^l)_1 \leq \dots \leq (\theta_l^T \mathbf{V}_i \mathbf{Q}_i^l)_M.
\end{align}
To preserve the order of the reference set elements across different samples, we focus on the effective permutation matrix $\mathbf{R}_i^l = \mathbf{Q}_i^l\left(\mathbf{P}^{l}\right)^T$, leading to the Monge displacement between $\mu^{l}$ and $\nu_i^{l}$, given by
\begin{align}\label{eq:monge_displacemnt_perm}
\mathbf{z}_i^l = \left[ (\theta_l^T \mathbf{V}_i \mathbf{R}_i^l )_1 - \theta_l^T \mathbf{u}_1 ~ \dots ~ (\theta_l^T \mathbf{V}_i \mathbf{R}_i^l )_M - \theta_l^T \mathbf{u}_M \right]^T \in \mathbb{R}^M.
\end{align}

Repeating the above procedure for all $L$ slices leads to the final embedding $e(\mathbf{V}_i; \mathbf{U}, \boldsymbol{\Theta}) =\left[(\mathbf{z}_i^1)^T ~ \dots ~ (\mathbf{z}_i^L)^T\right]^T\in\mathbb{R}^{LM}$, where we use $e(\cdot; \mathbf{U}, \boldsymbol{\Theta}): \mathbb{R}^{d \times m} \to \mathbb{R}^{LM}, \forall m\in\mathbb{N}$, to denote the entire permutation-invariant SWE pooling pipeline. The resulting embedding is ultimately fed to a prediction head $h(\cdot; \psi): \mathbb{R}^{LM} \to \mathcal{Y}$, parameterized by $\psi \in \Psi$, in order to make the final prediction, e.g., a feed-forward model (FFN). We use the shorthand notation $p:\mathcal{X} \to \mathcal{Y}$ to represent the end-to-end pipeline comprising the backbone, pooling, and prediction head, i.e.,
\vspace{-.03in}
\setlength{\belowdisplayskip}{4pt}
\setlength{\belowdisplayshortskip}{4pt}
\begin{align}
p(X_i;\phi, \psi, \mathbf{U}, \boldsymbol{\Theta}) \coloneqq h\bigg(e\Big(g(X_i; \phi); \mathbf{U}, \boldsymbol{\Theta}\Big); \psi\bigg).
\end{align}

\vspace{-.1in}
Letting $\ell: \mathcal{Y} \times \mathcal{Y} \to \mathbb{R}$ denote a loss function (e.g., the cross entropy loss), we can re-formulate the constrained learning problem~\eqref{eq:opt_constrained_relaxed} for the use case of constrained SWE pooling as
\vspace{-.03in}
\begin{subequations}\label{eq:opt_constrained_relaxed_swe}
\begin{alignat}{2}
&\min_{\phi\in\Phi, ~\psi\in\Psi, ~\mathbf{U}\in\mathbb{R}^{d \times M},~\boldsymbol{\Theta} \in (\mathbb{S}^{d-1})^L,~\mathbf{s}\in\mathbb{R}_+^L} &~~& \frac{1}{N} \sum_{i=1}^N \ell\Big(p(X_i;\phi, \psi, \mathbf{U}, \boldsymbol{\Theta}), y_i\Big) + \frac{\alpha}{2} \|\mathbf{s}\|_2^2, \label{eq:obj_relaxed_swe}        \\
&\qquad\qquad\qquad~~\text{s.t.} && \frac{1}{N} \sum_{i=1}^N \mathcal{D}_2(\mu^l, \nu_i^l; \theta_l) \leq \epsilon_l + s_l, \quad \forall l\in\{1,\dots,L\},\label{eq:swgg_constraint_relaxed_swe}
\end{alignat}
\end{subequations}
where on the LHS of~\eqref{eq:swgg_constraint_relaxed_swe}, the SWGG dissimilarities are averaged across the $N$ training samples (and their corresponding projected empirical distributions)\footnote{The SWGG constraints could alternatively be imposed \emph{per sample}. However, in that case, the number of constraints (as well as slack and dual variables) increases proportionally to the number of training samples, which could negatively impact the convergence of the primal-dual training algorithm.}. The primal-dual method in Algorithm~\ref{alg:pd} can then be used to solve this constrained learning problem, with the primal (stochastic) gradient descent updates now extended to the rest of the primal parameters, i.e., $\phi, \psi$, and $\mathbf{U}$, as well.

\subsection{Differentiability of Constraints with respect to Slicer Parameters}
\vspace{-.03in}

Combining~\eqref{eq:lagrangian},~\eqref{eq:slice_update}, and~\eqref{eq:opt_constrained_relaxed_swe}, we can expand the gradient descent step on the parameters of the $l$\textsuperscript{th} slicer, $l\in\{1,\dots,L\}$ as
\vspace{-.03in}
\begin{align}
\theta_l &\leftarrow \theta_l -\frac{\eta_{\boldsymbol{\Theta}}}{N} \sum_{i=1}^N \left[\frac{\partial \ell\Big(p(X_i;\phi, \psi, \mathbf{U}, \boldsymbol{\Theta}), y_i\Big)}{\partial {\theta_l}} + \lambda_l \frac{\partial \mathcal{D}_2(\mu^l, \nu_i^l; \theta_l)}{\partial {\theta_l}} \right].
\end{align}
Assuming the differentiability of the loss function and the main pipeline, the SWGG dissimilarities should also be differentiable with respect to the slicer parameters. For discrete distributions $\mu^{l}=\frac{1}{M}\sum_{j=1}^{M} \delta_{\theta_l^T \mathbf{u}_{j}}$ and $\nu_i^{l}=\frac{1}{M_i}\sum_{j=1}^{M_i} \delta_{\theta_l^T \mathbf{v}_{ij}}$, the SWGG dissimilarity in~\eqref{eq:swgg} can be simplified as
\vspace{-.03in}
\begin{align}\label{eq:swgg_discrete}
\mathcal{D}_2(\mu^l, \nu_i^l; \theta_l) = \left(\frac{1}{M_i}\sum_{j=1}^M\sum_{k=1}^{M_i}\|\mathbf{u}_{j}-\mathbf{v}_{ik}\|_2^2 (\mathbf{R}_i^l)_{jk}\right)^\frac{1}{2},
\end{align}
where $\mathbf{R}_i^l$ is the effective permutation matrix between the reference set and the input embedding set, as defined earlier based on the permutation matrices $\mathbf{P}^l$ and $\mathbf{Q}_i^l$ in~\eqref{eq:sort_perm}.

Now, observe that the only term in~\eqref{eq:swgg_discrete} that is a function of the slicer parameters is the permutation matrix $\mathbf{R}_i^l = \mathbf{Q}_i^l\left(\mathbf{P}^{l}\right)^T$
. However, the permutation matrices $\mathbf{P}^{l}$ and $\mathbf{Q}_i^l$ are derived based on the argsort operation and are not differentiable with respect to the elements of the vectors being sorted (i.e., $\theta_l^T \mathbf{U}$ and $\theta_l^T \mathbf{V}_i$, respectively). To resolve this issue, we propose to use the \emph{softsort} operation~\cite{prillo2020softsort,shahbazi2025espformer_arxiv} to make the permutation matrices differentiable with respect to the slicer parameters. More specifically, for a vector $\mathbf{x}\in\mathbb{R}^M$, we replace the hard permutation matrix with the following differentiable approximation,
\vspace{-.03in}
\begin{align}\label{eq:softsort}
    \mathbf{P}_\tau^d(\mathbf{x}) \coloneqq \mathsf{softmax}\left(\frac{-1}{\tau} \left\|(\mathsf{sort}(\mathbf{x})\mathds{1}_M^T- \mathds{1}_M\mathbf{x}^T)\right\|_2 \right),
\end{align}
where softmax is applied row-wise, $\tau>0$ is a temperature hyperparameter controlling the ``softness'' of the sorting operation, and $\mathds{1}_M$ denotes an $M$-dimensional vector with all entries equal to 1.

Note that using the approximation~\eqref{eq:softsort} for SWGG dissimilarities~\eqref{eq:swgg_discrete} when calculating the permutation matrix $\mathbf{R}_i^l = \mathbf{Q}_i^l\left(\mathbf{P}^{l}\right)^T$ increases the sorting computational complexity from $\mathcal{O}(M\log M)$ to $\mathcal{O}(M^2)$. However, this extra computational complexity is only necessary during the primal-dual training phase. Once the model is trained, the softsorting process is not needed during inference.

\section{Numerical Results}\label{sec:experiments}
\vspace{-.03in}

In this section, we present numerical results on the performance of the proposed method in three domains of images, point clouds, and protein sequences. We particularly show the benefits of SWGG-based constraints in learning informative slicing directions when pooling embeddings of pre-trained foundation models. 
Details on the experimental setup can be found in Appendix~\ref{appx:experiment_setup}.

\subsection{Image Classification with Vision Transformers}
\vspace{-.03in}

\setlength{\columnsep}{16pt}%
\begin{wrapfigure}[13]{r}{0.38\textwidth}
    \centering
    \vspace{-.3in}
    \includegraphics[ width=\linewidth]{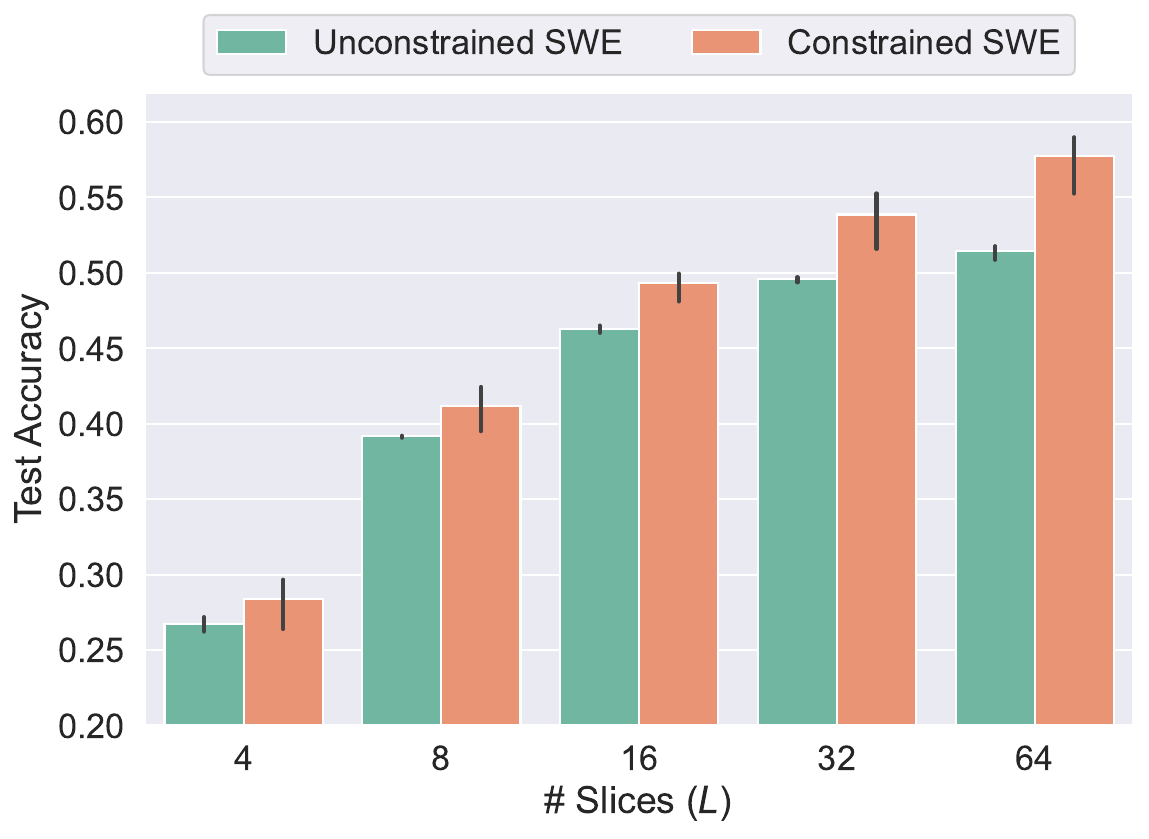}
    \caption{Classification accuracy of constrained vs.\ unconstrained SWE on Tiny ImageNet. Means and standard deviations are reported based on three runs.}
    \label{fig:Top1Flatten}
\end{wrapfigure}
We first consider the task of image classification using DeiT-Tiny \cite{touvron2021training}, a Vision Transformer (ViT) trained and fine-tuned on ImageNet1k \cite{deng2009imagenet} with 12 transformer layers and a classifier layer. We freeze the backbone transformer layers and train a classifier on the pooled embeddings 
using Tiny ImageNet~\cite{stanford2025tiny}, which contains 200 classes, and it poses a challenging task due to having a relatively small number of samples and a large number of classes. 
Figure~\ref{fig:Top1Flatten} compares the performance of constrained SWE with traditional SWE. We observe that the gain of constraining SWE over unconstrained SWE increases as $L$ grows.

Furthermore, Table~\ref{table:accuracy_layers} demonstrates the performance of constrained and unconstrained SWE (with $L=128$ slices) in earlier layers compared to global average pooling (GAP). Constrained SWE (C-SWE) performs better than GAP when the tokens are extracted after layer 6 and only slightly worse than GAP in later layers due to overfitting (stemming from larger embedding size). 
Observe that constrained SWE overfits much less than regular SWE, demonstrating the benefits of the constraints in improving the generalizability of SWE. Additional results can be found in Appendix~\ref{appx:addl_vit_results}.




\aboverulesep=0ex
\belowrulesep=0ex
\begin{table}[h!]
\scriptsize%
\centering%
\noindent\makebox[\textwidth]{%
\rowcolors{1}{Gray}{}%
\setlength{\tabcolsep}{2.8pt}%
\begin{tabular}{c|ccc|ccc|ccc}
\cmidrule[1.5pt]{1-10}
\cellcolor{Gray} & \multicolumn{3}{c|}{\textbf{Layer 6}} 
 & \multicolumn{3}{c|}{\textbf{Layer 9}} 
 & \multicolumn{3}{c}{\textbf{Layer 12}} \\
\cellcolor{Gray} \multirow{-2}{*}{\textbf{Pooling}} & \textbf{Train} & \textbf{Validation} & \textbf{Test} 
  & \textbf{Train} & \textbf{Validation} & \textbf{Test} 
  & \textbf{Train} & \textbf{Validation} & \textbf{Test} \\
\cmidrule[.5pt]{1-10}
C-SWE               & 60.08 (3.56) & 48.65 (0.80) & 49.00 (0.41)  &
                    73.76 (2.47) & 60.01 (0.90) & 59.48 (0.29) &  
                    71.15 (0.51) & 57.76 (0.42) & 57.87 (0.27) \\

SWE                 & 88.57 (0.07) & 42.31 (0.29) & 43.02 (0.77) & 
                    90.43 (0.03) & 55.44 (0.36) & 54.86 (0.26) & 
                    97.22 (0.12) & 53.88 (0.13) & 54.26 (0.52) \\

GAP                 & 54.10 (1.11) & 47.19 (0.07) & 47.88 (0.06) & 
                    68.17 (0.52) & 62.20 (0.03) & 62.00 (0.17) & 
                    66.78 (0.31) & 59.98 (0.02) & 60.19 (0.09) \\
\cmidrule[1.5pt]{1-10}
\end{tabular}}
\vspace{.1in}
\caption{Tiny ImageNet accuracies (mean (std) across three runs) of constrained SWE and unconstrained SWE (both with \(L=128\)), and GAP, using tokens extracted after layers 6, 9, and 12 of DeiT-Tiny.}
\label{table:accuracy_layers}
\end{table}

\vspace{-.2in}
\subsection{Point Cloud Classification with Point Cloud Transformers}
\vspace{-.03in}

\setlength{\columnsep}{16pt}%
\begin{wrapfigure}[12]{r}{0.38\textwidth}
    \centering
    \vspace{-.2in}
    \includegraphics[trim=0in 4.7in 0in 0in, clip, width=\linewidth]{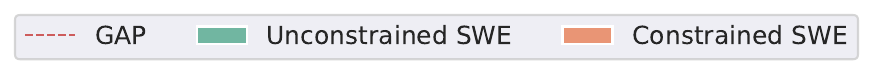}\vspace{.05in}
    \includegraphics[trim=0in 0in 0in 1.05in, clip, width=\linewidth]{figures/point_cloud.pdf}
    \vspace{-.2in}
    \caption{Test accuracies of PCT on ModelNet40 using unconstrained/constrained SWE and GAP. Means and standard deviations are reported based on three runs.}
    \label{fig:pointcloud}
    \vspace{-.3in}
\end{wrapfigure}

To evaluate the effectiveness of constrained SWE in point cloud classification, we conduct experiments using Point Cloud Transformers (PCT) \cite{guo2021pct} on the ModelNet40 dataset \cite{Zhirong15CVPR}, comprising 3D CAD models from 40 object categories. For each model, we sample 512 points to form a point cloud. We use a PCT backbone, pre-trained on the same ModelNet40 dataset, to map these point clouds to 256-dimensional embeddings. These embeddings are then aggregated using constrained SWE, unconstrained SWE, or GAP, followed by one layer of linear classification head. For constrained/unconstrained SWE, the classification task is performed across varying numbers of slices $L=4, 8, 16, 32, 64$ based on a reference size $M=512$, equal to the input size. As shown in Figure \ref{fig:pointcloud}, all configurations using SWE outperform GAP. SWE's performance improves with an increasing number of slices, and constrained SWE consistently outperforms unconstrained SWE across all numbers of slices, with particularly notable gains at lower slice numbers.

\subsection{Subcellular Localization with Protein Language Models}

We finally consider the task of subcellular localization of proteins, whose goal is to determine which compartment of the cell a protein localizes in~\cite{almagro2017deeploc,10.1093/bioadv/vbab035,li2024feature}. This task is formulated as a 10-class classification problem, with the input samples being a set of protein primary amino acid sequences. In order to map these protein sequences to high-dimensional embeddings, we leverage protein language models (PLMs) that have been trained on massive protein sequence databases using a self-supervised masked language modeling objective~\cite{weissenow2025protein,xiao2025protein,wang2025comprehensive}. In particular, we use four model architectures from the ESM-2 family of PLMs trained on the UniRef50 database~\cite{suzek2015uniref}, with sizes ranging from 8 to 650 million parameters~\cite{lin2023evolutionary}. We use each of the PLMs to derive token-level embeddings of a given protein sequence, aggregate them using constrained and unconstrained SWE to derive a protein-level embedding, and feed the aggregated embedding to a linear classifier head to derive class probabilities.

\begin{figure}[h]
\centering
\includegraphics[width=\linewidth]{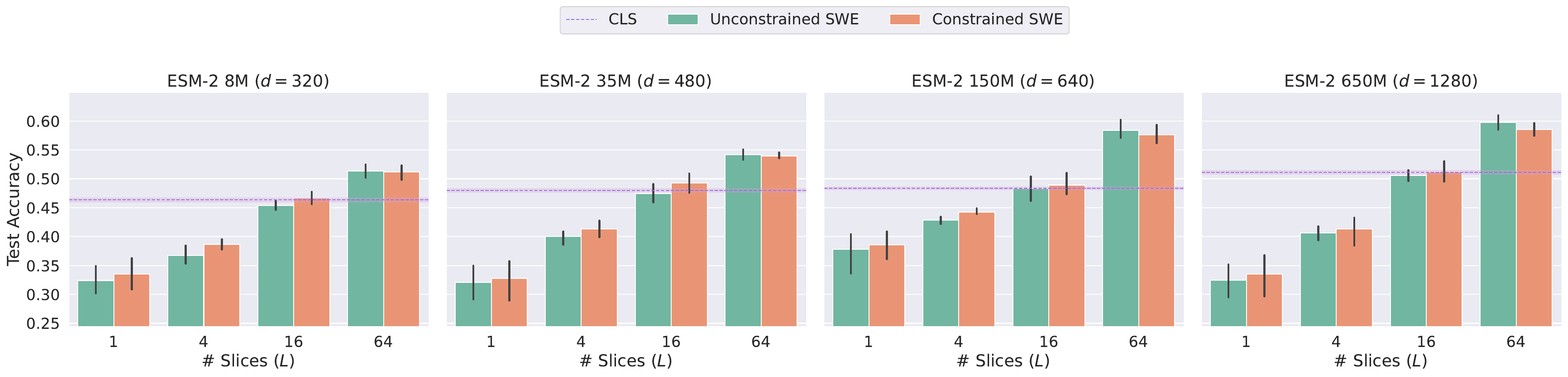}
\caption{Test accuracy of the proposed method as compared to unconstrained SWE and CLS token embedding on the subcellular localization task across four ESM-2 protein language models (PLMs)~\cite{lin2023evolutionary} with 8, 35, 150, and 650 million parameters (from left to right). Means and standard deviations are reported based on five runs.}
\label{fig:plm_scl_swe_cswe_cls}
\end{figure}

Figure~\ref{fig:plm_scl_swe_cswe_cls} compares the performance of constrained SWE with traditional SWE and the CLS token embedding. As the figure shows, the classification performance generally improves with more slices and more expressive PLM architectures. The gains of constrained SWE over traditional SWE are most significant for fewer numbers of slices. As the number of slices increases, the performance gains of constrained SWE fade away as the number of slices increases, potentially due to the constrained optimization problem becoming infeasible. Quite interestingly, the CLS token embedding performance is approximately equivalent to $L=16$ slices of constrained SWE across all four PLMs. The protein sequences differ in length, and all of these experiments were conducted with $M=100$ reference points. Larger hyperparameter search spaces, especially over $M$ and $\{\epsilon_l\}_{l=1}^L$, may be used to improve the performance of constrained SWE for larger numbers of slices. We provide detailed results on the evolution of SWGG levels, as well as slack and dual variables, in Appendix~\ref{appx:convergence_trends_plm}.

\section{Discussion and Concluding Remarks}\label{sec:conc}

We proposed a constrained learning framework for optimizing slicing directions in sliced Wasserstein embeddings (SWE), enforcing that the resulting one-dimensional transport plans remain meaningful in the original high-dimensional space. Using a relaxed primal-dual formulation, our method selects more informative slices, enabling lower embedding dimensionality while preserving or improving performance. A key advantage of SWE is that the embedding size grows linearly with the number of slices. By learning higher-quality slices, our method achieves stronger performance with fewer slices, reducing computational cost and improving efficiency in downstream tasks. 

Several limitations suggest directions for future work. Currently, our embeddings are flattened across slices, but more expressive aggregation strategies, such as using dual or slack variables as slice-wise importance weights, may improve performance. Our framework also supports additional constraint types, such as orthogonality or Max-SW-style constraints, which could further enhance slice heterogeneity or informativeness. Finally, hybrid approaches that balance dissimilarity maximization after slicing and SWGG alignment before slicing may lead to stronger generalization capabilities.








\appendix

\section{Extension of SWE to Different Numbers of Tokens}\label{appx:interp}

In order to derive the embedding (corresponding to the $l$\textsuperscript{th} slice, $l\in\{1,\dots,L\}$) for a sliced empirical distribution $\nu_i^{l}=\frac{1}{M_i}\sum_{j=1}^{M_i} \delta_{\theta_l^T \mathbf{v}_{ij}}$ with $M_i \neq M$ tokens, we first derive the permutation matrices $\mathbf{Q}_i^l \in \mathbb{S}_{M_i}$ and $\mathbf{P}^l \in \mathbb{S}_M$ for $\nu_i^{l}$ and the sliced reference distribution $\mu^{l}$, respectively. We then replace the effective permutation matrix $\mathbf{R}_i^l$ in ~\eqref{eq:monge_displacemnt_perm} with $\mathbf{R}_i^l = \mathbf{Q}_i^l I_i \left(\mathbf{P}^{l}\right)^T$, where $I_i \in \mathbb{R}^{M_i \times M}$ is a linear interpolation matrix, whose entries are given by
\begin{equation}
    I_i[j,m]\coloneqq\begin{cases}
        1 - \chi_j,&\text{if } m=m_j;\\
        \chi_j,&\text{if } m=m_j+1;\\
        0,&\text{o.w.}
    \end{cases}
\end{equation}
where $m_j=\lfloor (j-1)\frac{M-1}{M_i-1} \rfloor + 1$ and $\chi_j = (j-1)\frac{M-1}{M_i-1} + 1 - m_j$.

\section{Experimental Settings}\label{appx:experiment_setup}

\subsection{Hyperparameter Search}

Table~\ref{table:hyperparameters} shows the hyperparameter grids that we used for optimizing the performance of constrained SWE on the three tasks studied in Section~\ref{sec:experiments}. In all experiments, the upper bounds are taken to be the same across all slices, i.e., $\epsilon_l=\epsilon, \forall l\in\{1,\dots,L\}$. Furthermore, in all three tasks, during training, the backbone is kept frozen (i.e., $\phi$ is removed from the primal optimization variables in~\eqref{eq:opt_constrained_relaxed_swe}), and only the pooling layer and classification head are optimized.

\aboverulesep=0ex
\belowrulesep=0ex
\begin{table}[h!]
\scriptsize%
\centering%
\noindent\makebox[\textwidth]{%
\rowcolors{1}{Gray}{}%
\setlength{\tabcolsep}{6pt}%
\begin{tabular}{l|ccc}
\cmidrule[1.5pt]{1-4}
\textbf{Hyperparameter} 
  & \textbf{Image Classification} 
  & \textbf{Point Cloud Classification} 
  & \textbf{Subcellular Localization} \\
\hhline{|*4{-}|}\hhline{|*4{-}|}
$\epsilon$ (constraint upper bound)     
  & $\{11,15,17,18,19,20,21,22,24\}$ 
  & $\{1, 3.5, 5, 7\}$
  & $\{5,10\}$ \\

$\alpha$ (slack norm coefficient)       
  & $\{0.1,0.5,1\}$ 
  & $\{0.1,1\}$ 
  & $\{0.1,1\}$ \\

$\eta_{\boldsymbol{\lambda}}$ (dual learning rate) 
  & $\{0.001,0.01\}$ 
  & $\{0.001,0.01\}$  
  & $\{0.001,0.01\}$ \\

$\eta_{s}$ (slack learning rate)        
  & $\{0.001,0.01\}$  
  & $\{0.001\}$   
  & $\{0.01\}$ \\

$\tau$ (softsort temperature)           
  & $\{0.001,0.01\}$ 
  & $\{0.001,0.01\}$
  & $\{0.001,0.01\}$ \\
\cmidrule[1.5pt]{1-4}
\end{tabular}}
\vspace{.1in}
\caption{Grid of hyperparameters used for the numerical experiments.}
\label{table:hyperparameters}
\end{table}

\subsection{Image Classification} 

\aboverulesep=0ex
\belowrulesep=0ex
\begin{table}[h!]
\scriptsize
\centering
\rowcolors{1}{Gray}{}%
\setlength{\tabcolsep}{6pt}%
\begin{tabular}{l*{9}{l}}
\cmidrule[1.5pt]{1-10}
\textbf{Hyperparameter}   & \textbf{Batch size} & \textbf{Epochs} & \textbf{Primal Learning Rate ($\eta_{p}$)} & \textbf{$\eta_{s}$} & \textbf{$\eta_{\lambda}$} & \textbf{$\alpha$} & \textbf{$\epsilon$} & \textbf{$\tau$} & \textbf{$M$} \\
\hhline{|*{10}{-}|}\hhline{|*{10}{-}|}
\textbf{Chosen Value}     & 1024                & 80              & 0.001                & 0.001                & 0.001                     & 0.1               & 21                  & 0.01             & 196           \\
\cmidrule[1.5pt]{1-10}
\end{tabular}
\vspace{.1in}
\caption{Selected hyperparameters for the DeiT-Tiny experiments.}
\label{table:hyperparams-img-selected}
\end{table}

Table~\ref{table:hyperparams-img-selected} shows the hyperparameters used for the image classification experiments. We use the Adam optimizer~\cite{adam} for training the pooling, classifier, and slack parameters. For these parameters, we use a StepLR scheduler, with the primal and slack learning rates reducing to 10\% of their initial value at epoch 60. For each run, the ``best'' model—whose accuracies we record—is the one with the highest validation accuracy for correctly classifying an image (i.e., top-1 accuracy).

For hyperparameter tuning, our constrained-learning approach motivated choosing an $\epsilon$ that was lower than the average unconstrained SWGG dissimilarity per batch. Empirically, the mean SWGG dissimilarity across all slices was around $26$–$28$ (for all $L$). As a result, we conducted a sweep of epsilon below this value to find the optimal constraint bound, which generalized across slices.

We set the size of our reference set ($M$) equal to the number of tokens (i.e., the number of patches) per image, so no interpolation is required when computing Monge couplings between the sliced sample’s 1D distributions and the sliced reference set’s 1D distributions.

We use the original validation set of Tiny ImageNet as our test set. Additionally, our train and validation sets are from a 90-10 split on the original training data.

Table~\ref{table:compute-resources-image} shows the compute resources we used for the image classification experiments.

\aboverulesep=0ex
\belowrulesep=0ex
\begin{table}[H]
\scriptsize%
\centering%
\noindent\makebox[\textwidth]{%
\rowcolors{1}{Gray}{}%
\setlength{\tabcolsep}{6pt}%
\begin{tabular}{l|l}
\cmidrule[1.5pt]{1-2}
\textbf{Resource / Metric}            & \textbf{Details} \\ 
\hhline{|*2{-}|}\hhline{|*2{-}|}
Compute environment                   & Internal GPU cluster \\
NVIDIA GPU types                             & RTX 6000 Ada; H200 \\
Experiment variants                   & $L$ = $\{4,8,16,32\}$, CLS on RTX 6000 Ada; $L=64$ on H200 \\
Runtime per run                       & 3–7 hours for $L\leq32$; 6–12 hours for $L=64$ \\
Memory requirement per run            & $\leq 40\,$GB for $L\leq32$; $\approx 80\,$GB for $L=64$ \\
Hyperparameter-tuning             & $\approx$ 30 runs on RTX 6000 Ada \\
Alternative SWE embedding size reduction methods         & $\approx$ 10 runs on RTX 6000 Ada \\
\cmidrule[1.5pt]{1-2}
\end{tabular}}
\vspace{.1in}
\caption{Compute resources for DeiT-Tiny experiments.}
\label{table:compute-resources-image}
\end{table}

\subsection{Point Cloud Classification}
We use the original train-test split for ModelNet40 \cite{Zhirong15CVPR} with 9840 training samples and 2468 test samples, and then 20\% of the training data is extracted to form a validation set for the purpose of hyperparameter tuning. 

Table~\ref{table:hyperparams-pc-selected} shows the hyperparameters used for the point cloud classification experiments.

\aboverulesep=0ex
\belowrulesep=0ex
\begin{table}[h!]
\scriptsize
\centering
\rowcolors{1}{Gray}{}%
\setlength{\tabcolsep}{6pt}%
\begin{tabular}{l*{9}{l}}
\cmidrule[1.5pt]{1-10}
\textbf{Hyperparameter}   & \textbf{Batch size} & \textbf{Epochs} & \textbf{$\eta_{p}$} & \textbf{$\eta_{s}$} & \textbf{$\eta_{\lambda}$} & \textbf{$\alpha$} & \textbf{$\epsilon$} & \textbf{$\tau$} & \textbf{$M$} \\
\hhline{|*{10}{-}|}\hhline{|*{10}{-}|}
\textbf{Chosen Value}     & 128 for $L = \{4,8,16,32\}$, 64 for $L=64$ & 200            & 0.001                & 0.001                & 0.001                     & 1                 & 7                   & 0.001             & 512           \\
\cmidrule[1.5pt]{1-10}
\end{tabular}
\vspace{.1in}
\caption{Selected hyperparameters for the point cloud classification experiments.}
\label{table:hyperparams-pc-selected}
\end{table}

The pooling layers and the linear classification heads are trained using an Adam optimizer with a StepLR scheduler. The primal and slack learning rates are decayed by 50\% every 50 epochs.

Table~\ref{table:compute-resources-point-cloud} shows the compute resources we used for the point cloud classification experiments.

\aboverulesep=0ex
\belowrulesep=0ex
\begin{table}[H]
\scriptsize%
\centering%
\noindent\makebox[\textwidth]{%
\rowcolors{1}{Gray}{}%
\setlength{\tabcolsep}{6pt}%
\begin{tabular}{l|l}
\cmidrule[1.5pt]{1-2}
\textbf{Resource / Metric}            & \textbf{Details} \\ 
\hhline{|*2{-}|}\hhline{|*2{-}|}
Compute environment                   & Internal GPU cluster \\
NVIDIA GPU types                             & RTX A5000 \\
Runtime per run                       & 1–12 hours \\
Memory requirement per run            & $\leq 30\,$GB  \\
\cmidrule[1.5pt]{1-2}
\end{tabular}}
\vspace{.1in}
\caption{Compute resources for the point cloud classification experiments.}
\label{table:compute-resources-point-cloud}
\end{table}

\subsection{Subcellular Localization} We use the AdamW optimizer~\cite{loshchilov2018decoupled} for training the slicer, classifier, and slack parameters. The slicer and classifier learning rate is initially set to $10^{-4}$ and varied using a cosine annealing scheduler with warm restarts every 10 epochs~\cite{loshchilov2017sgdr}. We decrease the slack and dual learning rates by 5\% every epoch. We use a batch size of 32 and train each model for 50 epochs. For each PLM type, the model checkpoint with the hyperparameter combination and training epoch that leads to the highest validation accuracy is saved and evaluated on the test set.

Table~\ref{table:hyperparams-plm-selected} shows the hyperparameters used for the subcellular localization experiments. Moreover, Table~\ref{table:compute-resources-plm} shows the compute resources we used for these experiments.

\aboverulesep=0ex
\belowrulesep=0ex
\begin{table}[h!]
\scriptsize%
\centering%
\noindent\makebox[\textwidth]{%
\rowcolors{1}{Gray}{}%
\setlength{\tabcolsep}{6pt}%
\begin{tabular}{l|cccc}
\cmidrule[1.5pt]{1-5}
\textbf{\# Slices ($L$)} 
  & \textbf{ESM-2 8M} 
  & \textbf{ESM-2 35M} 
  & \textbf{ESM-2 150M}
  & \textbf{ESM-2 650M} \\
\hhline{|*5{-}|}\hhline{|*5{-}|}
1    
  & $(5, 0.001, 1, 0.01)$ 
  & $(10, 0.001, 1, 0.01)$
  & $(10, 0.001, 1, 0.01)$
  & $(5, 0.001, 0.1, 0.01)$ \\
4       
  & $(5, 0.001, 1, 0.001)$ 
  & $(5, 0.001, 0.1, 0.001)$ 
  & $(10, 0.001, 1, 0.001)$
  & $(10, 0.001, 0.1, 0.001)$  \\
16
  & $(5, 0.001, 0.1, 0.001)$ 
  & $(10, 0.001, 1, 0.001)$  
  & $(10, 0.001, 1, 0.001)$
  & $(10, 0.001, 0.1, 0.001)$  \\
64        
  & $(10, 0.001, 0.1, 0.01)$  
  & $(10, 0.001, 0.1, 0.01)$   
  & $(10, 0.001, 0.1, 0.01)$
  & $(10, 0.001, 0.1, 0.01)$  \\
\cmidrule[1.5pt]{1-5}
\end{tabular}}
\vspace{.1in}
\caption{Hyperparameters selected for the subcellular localization experiments across different numbers of slices and PLM architectures. Each hyperparameter tuple denotes the values for $\epsilon$, $\eta_{\boldsymbol{\lambda}}$, $\alpha$, and $\tau$, respectively.}
\label{table:hyperparams-plm-selected}
\end{table}

\aboverulesep=0ex
\belowrulesep=0ex
\begin{table}[H]
\scriptsize%
\centering%
\noindent\makebox[\textwidth]{%
\rowcolors{1}{Gray}{}%
\setlength{\tabcolsep}{6pt}%
\begin{tabular}{l|l}
\cmidrule[1.5pt]{1-2}
\textbf{Resource / Metric}            & \textbf{Details} \\ 
\hhline{|*2{-}|}\hhline{|*2{-}|}
Compute environment                   & Internal GPU cluster \\
NVIDIA GPU types                             & Tesla P100; RTX 2080Ti; RTX A5000; RTX 5000 Ada; RTX 6000 Ada; H200 \\
Runtime per run                       & 1–12 hours \\
Memory requirement per run            & $\leq 30\,$GB  \\
\cmidrule[1.5pt]{1-2}
\end{tabular}}
\vspace{.1in}
\caption{Compute resources for the subcellular localization experiments.}
\label{table:compute-resources-plm}
\end{table}

\section{Additional Image Classification Results}\label{appx:addl_vit_results}

\begin{figure}[H]
  \centering

  \begin{subfigure}[t]{0.45\textwidth}
    \centering
    \includegraphics[width=\linewidth]{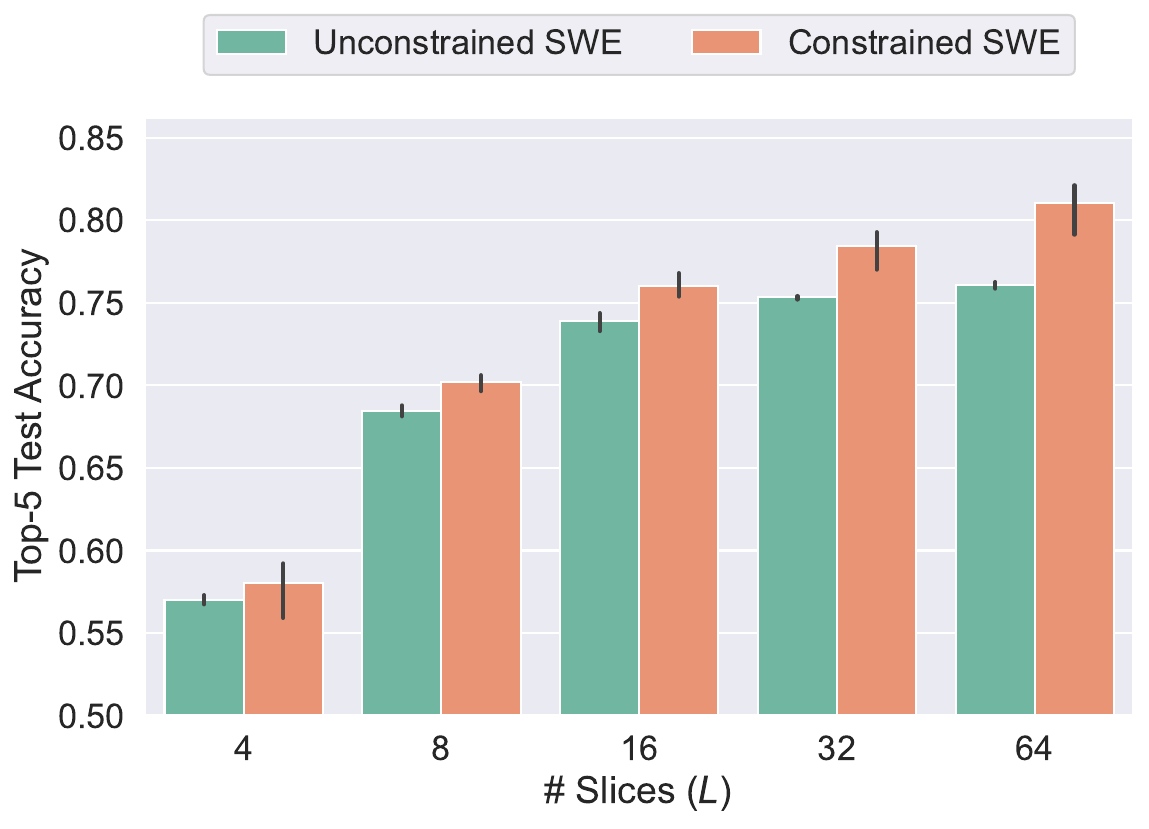}
    \caption{Means and standard deviations are reported based on three runs.}
    \label{fig:Top5Flatten}
  \end{subfigure}\hfill
  \begin{subfigure}[t]{0.45\textwidth}
    \centering
    \includegraphics[width=\linewidth]{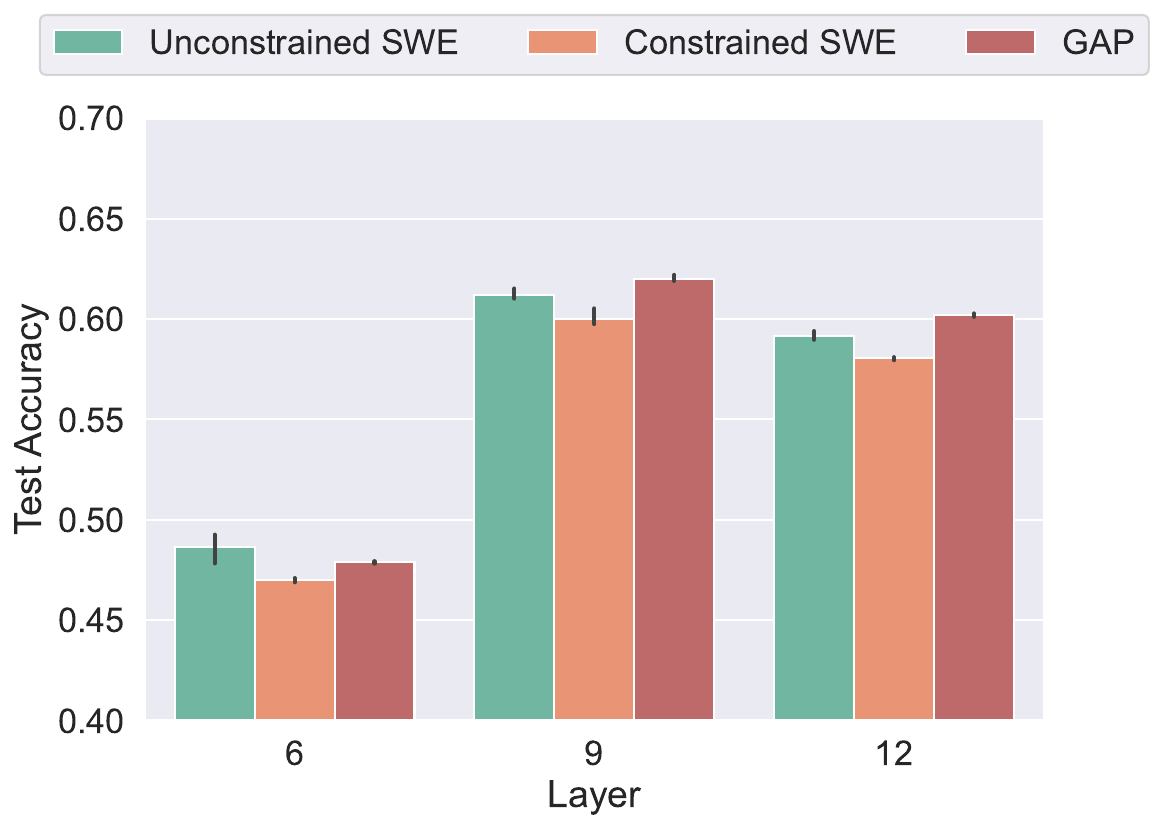}
    \caption{Means and standard deviations are reported based on three runs.}
    \label{fig:MeanCompMapL128}
  \end{subfigure}

  \caption{(a) Top-5 classification accuracy of constrained vs. unconstrained SWE on Tiny ImageNet, and (b) top-1 test accuracy of constrained SWE and unconstrained SWE (both with $L=128$), and GAP using tokens extracted after layers 6, 9, and 12 of DeiT-Tiny; constrained and unconstrained SWE embeddings were reduced to $L$ dimensions using a learnable mapping.}
  \label{fig:add_results}
\end{figure}

Figure~\ref{fig:Top5Flatten} shows the top-5 image classification accuracy comparison between constrained and unconstrained SWE. Moreover, to address the potential issue of lacking fair comparison, instead of flattening the SWE and constrained SWE embedding, we use an $M\to1$ learnable mapping to compress the embedding to dimension $L$, which results in a classification layer with only $L\times 200$ parameters. For $L=128$ specifically, the classification layer has approximately 1.5 times fewer parameters than that of a model using GAP, and the effects can be seen in Figure \ref{fig:MeanCompMapL128}. Other approaches to reduce the embedding size were tested, from taking the mean across dimension $L$ or $M$ to using a learnable mapping $L\to1$. Ultimately, mapping to an $L$-dimensional embedding performed the best.

\section{Evolution of SWGG Levels and Slack and Dual Variables}\label{appx:convergence_trends_plm}

\begin{figure}[h]
\centering
\includegraphics[width=\linewidth]{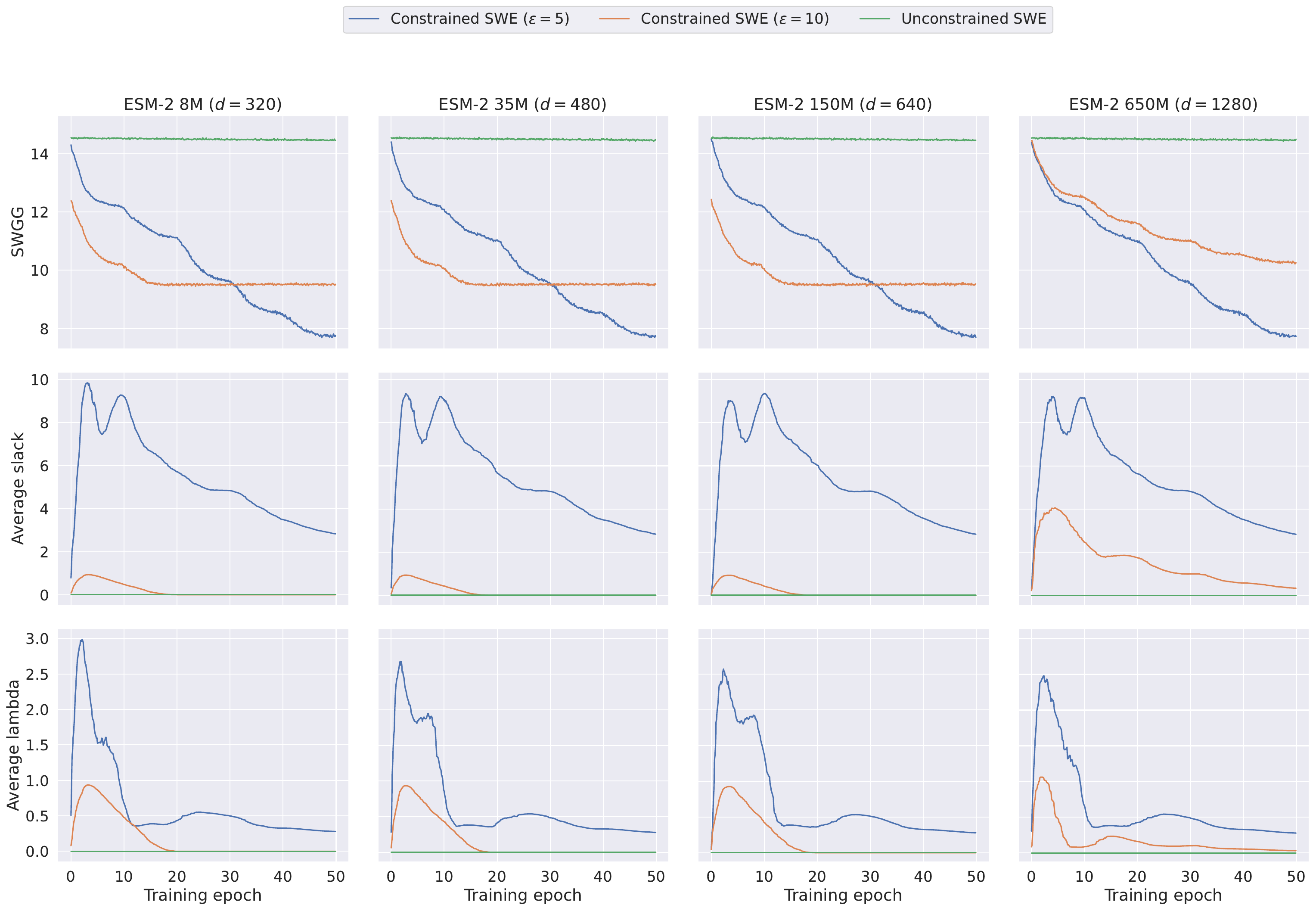}
\caption{The evolution of SWGG levels, slack variables, and dual variables in the subcellular localization task with $L=16$ slices across the four ESM-2 PLMs.}
\label{fig:convergence_trends_plm}
\end{figure}

Figure~\ref{fig:convergence_trends_plm} illustrates how the SWGG dissimilarity levels, as well as the slack variables and dual variables, evolve during the course of training for constrained and unconstrained SWE in the subcellular localization task with $L=16$ slices. As the figure demonstrates, while the SWGG level for unconstrained SWE remains virtually constant, our proposed method finds slicing directions that find the right balance between minimizing the main classification objective and reducing SWGG levels. The smaller the constraint upper bound $\epsilon$ is, the more challenging it is for the slices to satisfy the constraints, leading to elevated slack and dual variables.

\section{Licenses for Models and Datasets}

\textbf{Image Classification.} The Tiny-ImageNet dataset, a subset of ImageNet \cite{deng2009imagenet}, was made available under the ImageNet Terms of Use (non-commercial research and educational license only). We also used the DeiT-Tiny model released under the Apache License 2.0.

Relevant data and models used can be found at: \\
\url{http://cs231n.stanford.edu/tiny-imagenet-200.zip}\\
\url{https://github.com/facebookresearch/deit}

\textbf{Point Cloud Classification.} The ModelNet40 dataset was downloaded under Princeton University's non-commercial academic research terms. We used the Point Cloud Transformer model released under the MIT License.

Relevant data and models used can be found at: \\
\url{https://huggingface.co/datasets/Msun/modelnet40/resolve/main/modelnet40_ply_hdf5_2048.zip} \\
\url{https://github.com/Strawberry-Eat-Mango/PCT_Pytorch}

\textbf{Subcellular Localization.} Protein sequence data were obtained from Zenodo under the Creative Commons Zero v1.0 Universal. We used the ESM-2 pretrained models from Facebook Research, which are released under the MIT License. 

Relevant data and models used can be found at: \\
\url{https://zenodo.org/records/10631963} \\
\url{https://github.com/facebookresearch/esm}


\bibliographystyle{abbrvnat}
\bibliography{references}

\end{document}